\pdfoutput=1
\documentclass[sigconf]{acmart}
\usepackage{soul}
\usepackage{amsmath}
\usepackage{subfig}

\AtBeginDocument{%
  \providecommand\BibTeX{{%
    \normalfont B\kern-0.5em{\scshape i\kern-0.25em b}\kern-0.8em\TeX}}}

\acmConference
\acmPrice
\acmISBN

\renewcommand\footnotetextcopyrightpermission[1]{} 
\begin{document}

\title{Evaluation of Embedding Models for Automatic Extraction and Classification of Acknowledged Entities in Scientific Documents}

\author{Nina Smirnova}
\affiliation{%
  \institution{GESIS –- Leibniz Institute for the Social Sciences}
  \city{Cologne}
  \country{Germany}}
\email{nina.smirnova@gesis.org}

\author{Philipp Mayr}
\affiliation{%
  \institution{GESIS –- Leibniz Institute for the Social Sciences}
  \city{Cologne}
  \country{Germany}}
\email{philipp.mayr@gesis.org}

\renewcommand{\shortauthors}{Smirnova \& Mayr}

\begin{abstract} 
Acknowledgments in scientific papers may give an insight into aspects of the scientific community, such as reward systems, collaboration patterns, and hidden research trends. The aim of the paper is to evaluate the performance of different embedding models for the task of automatic extraction and classification of acknowledged entities from the acknowledgment text in scientific papers. We trained and implemented a named entity recognition (NER) task using the Flair NLP-framework. The training was conducted using three default Flair NER models with two differently-sized corpora. The Flair Embeddings model trained on the larger training corpus showed the best accuracy of 0.77. Our model is able to recognize six entity types: funding agency, grant number, individuals, university, corporation and miscellaneous. The model works more precise for some entity types than the others, thus, individuals and grant numbers showed very good F1-Score over 0.9. Most of the previous works on acknowledgement analysis were limited by the manual evaluation of data and therefore by the amount of processed data. This model can be applied for the comprehensive analysis of the acknowledgement texts and may potentially make a great contribution to the field of automated acknowledgement analysis.
\end{abstract}

\ccsdesc[500]{Applied computing}
\ccsdesc[500]{Applied computing~Document management and text processing}

\keywords{NLP, NER, Web of Science, Acknowledgement, Text Mining, Flair NLP-Framework}

\maketitle

\section{Introduction}

Acknowledgements in scientific papers are short texts where the author(s) \textit{“identify those who made special intellectual or technical contribution to a study that are not sufficient to qualify them for authorship”} \cite[p.~1511]{kassirer_authorship_1991}. Cronin and Weaver \cite{cronin_praxis_1995} ascribe an acknowledgment alongside authorship and citedness to measures of a researcher's scholarly performance: a feature that reflects the researcher’s productivity and impact. 
Giles and Councill \cite{giles_who_2004} argue that acknowledgments to individuals, in the same way as citations, may be used as a metric to measure an individual’s intellectual contribution to scientific work. Acknowledgements of financial support are interesting in terms of evaluating the influence of funding agencies on academic research. Acknowledgments of technical and instrumental support may reveal \textit{“indirect contributions of research laboratories and universities to research activities”} \cite[p.~17599]{giles_who_2004}. 

The analysis of acknowledgments is particularly interesting as acknowledgments may give an insight into aspects of the scientific community, such as reward systems, collaboration patterns, and hidden research trends. From the linguistic point of view, acknowledgements are unstructured text data, which through automatic analysis poses research and methodological problems like data cleaning, choosing the right tokenization method, and whether and how word embeddings may enhance their automatic analysis.

To our knowledge, previous works on automatic acknowledgment analysis were mostly concerned with the extraction of funding organizations and grant numbers \cite{alexandera_this_2021,kayal-etal-2017-tagging} or classification of acknowledgement texts \cite{song_kang_timakum_zhang}. 
Furthermore, large bibliographic databases such as Web of Science (WoS)\footnote{\url{http://wokinfo.com/products_tools/multidisciplinary/webofscience/fundingsearch/}} and Scopus selectively index only funding information, i.e., names of funding organizations and grant identification numbers. Consequently, we want to extend that to other types of acknowledged entities. 

The aim of the present paper is to evaluate the performance of existing embedding models for the task of automatic extraction and classification of acknowledged entities from the acknowledgment text in scientific papers. 
The Flair - an open-source Natural Language Processing (NLP) Framework \cite{akbik_flair_2019} is used in our study for creating a tool for extraction of acknowledged entities because this library is easily customizable. It offers the possibility of creating a customized Named Entity Recognition (NER) tagger, which can be used for processing and analyzing the acknowledgement texts. Furthermore, Flair has shown better accuracy for NER tasks using pre-trained datasets in comparison with many other open source NLP tools\footnote{\url{https://github.com/flairNLP/flair}}. 

We trained and implemented a NER task using three default Flair NER models with two differently-sized corpora\footnote{The release 0.9 (\url{https://github.com/flairNLP/flair/releases/tag/v0.9}) was used in the present research. All the descriptions of the Flair framework features refer to the 0.9 release.}. Models were trained to recognize six types of acknowledged entities: funding agency, grant number, individuals, university, corporation and miscellaneous. The model with the best accuracy can be applied for the comprehensive analysis of the acknowledgement texts.
We performed an additional training with altered training parameters or altered training corpora (Section~\ref{sec:add_train}).
Most of the previous works on acknowledgement analysis were limited by the manual evaluation of data and therefore by the amount of processed data \cite{giles_who_2004, paul_hus_all_2017,paul_hus_des_2019, Mccain2017}. Furthermore, Thomer and Weber \cite{thomer_weber_2014} argue that using of named entities can benefit the process of manual document classification and evaluation of the data. Therefore, a model, which is capable of extracting and classification of different entity types may potentially make a great contribution to the field of automated acknowledgement analysis.  

\subsection*{Research questions}
In this paper, we address the following research questions:
\begin{itemize}
    \item \textbf{RQ1:} Which of the Flair default NER models is more suitable for the defined task of the extraction and classification of acknowledged entities from scientific acknowledgements?
    \item \textbf{RQ2:} How does the training corpus size impact the training accuracy for different NER models? 
\end{itemize}

\section{Background and Related work}\label{sec:background}
Named Entity Recognition (NER) is a form of NLP, which aims to extract named entities from an unstructured text and classify them into predefined categories. A named entity is a real-world object that is important for understanding the text \cite{deepai_named-entity_2019}. As a rule NER tasks require training data, i.e., a particular dataset or corpus, which is usually divided into several datasets:  training set, test set and validation set. NER models require corpora with semantic annotation, i.e., metadata about concepts attached to the unstructured text data \cite{knowledge_hub_what_nodate}. The annotation process is crucial as insufficient or redundant metadata can slow down and bias a learning process \cite[Chapter~1]{pustejovsky_natural_2012}.

We are aware of several works on automated information extraction from acknowledgements. Giles and Councill \cite{giles_who_2004} developed an automated method for acknowledgment extraction and analysis using regular expressions and the support vector machines (SVM) classification algorithm. Computer science research papers from the CiteSeer digital library were used as the data source. Extracted entities were analysed and manually assigned to the following four categories: funding agencies, corporations, universities, and individuals.

Thomer and Weber \cite{thomer_weber_2014} used the 4 class Stanford Entity Recognizer \cite{finkel-etal-2005-incorporating} to extract persons, locations, organizations and miscellaneous entities from the collection of the bioinformatics texts from PubMed Central's Open Access corpus. Aim of the study was to determine an approach to \textit{"increase the speed of ... classification without sacrificing accuracy, nor reliability"} \cite[p.~1134]{thomer_weber_2014}. 

Kayal et al. \cite{kayal-etal-2017-tagging} introduced a method for extraction of funding organizations and grants from acknowledgement texts using a combination of sequential learning models: conditional random fields (CRF), hidden markov models (HMM) and maximum entropy models (MaxEnt). The final model contained pooling outputs of the single used models. 

Alexandera and de Vries \cite{alexandera_this_2021} proposed AckNER, a tool for financial information extraction from the funding or acknowledgment section of a research article. AckNER works with the use of dependency parse trees and regular expressions and is able to extract names of the organisations, projects, programs and funds, as also numbers of contracts and grants \footnote{AckNER showed better performance as Flair, but is specifically designed to recognize two types of acknowledged entities \cite{alexandera_this_2021}, which was insufficient for the present project.}. 

\subsection*{The Flair NLP Framework}\label{subsec:flair}

Flair is an open-sourced NLP framework built on PyTorch \cite{paszke_pytorch_2019}, which is an open source machine learning library. \textit{“The core idea of the framework is to present a simple, unified interface for conceptually very different types of word and document embeddings”} \cite[p.~54]{akbik_flair_2019}. Flair has three default training algorithms for NER which were used for primary training in the present research: a) NER Model with Flair Embeddings (later on Flair Embeddings) \cite{akbik_contextual_2018}, b) NER Model with Transformers (later on Transformers) \cite{schweter_flert_2020}, and c) Zero-shot NER with TARS (later on TARS) \cite{halder_task-aware_2020}.

The Flair Embeddings model uses stacked embeddings, i.e., a combination of contextual string embeddings with a static embeddings model. Contextual string embeddings is a new character based contextual string embeddings method proposed by Akbik et al. \cite{akbik_contextual_2018}. This approach will generate different embeddings for the same word depending on its context. Stacked embedding is an important Flair feature, as a combination of different embeddings might bring better results than their separate uses \cite{akbik_flair_2019}. 

The Transformers model or FLERT-extension (document-level features for NER) is a set of settings to perform a NER on document level using fine-tuning and feature-based LSTM-CRF with the multilingual XML-RoBERTa transformer model  \cite{schweter_flert_2020}. 

The TARS (task-aware representation of sentences) is a transformer-based model, which allows performing training without any training data (zero-shot learning) or with a small dataset (few-short learning) \cite{halder_task-aware_2020}. The TARS approach differs from the traditional transfer learning approach in a way that the TARS model also considers semantic information captured in the class labels themselves. For example, class labels like \textit{food} or \textit{sport} already carry semantic information \cite[p.~2]{halder_task-aware_2020}. 

\section{Method}\label{sec:usecase}

\begin{figure}[h!]
\centering
  \includegraphics[width=0.5\textwidth]{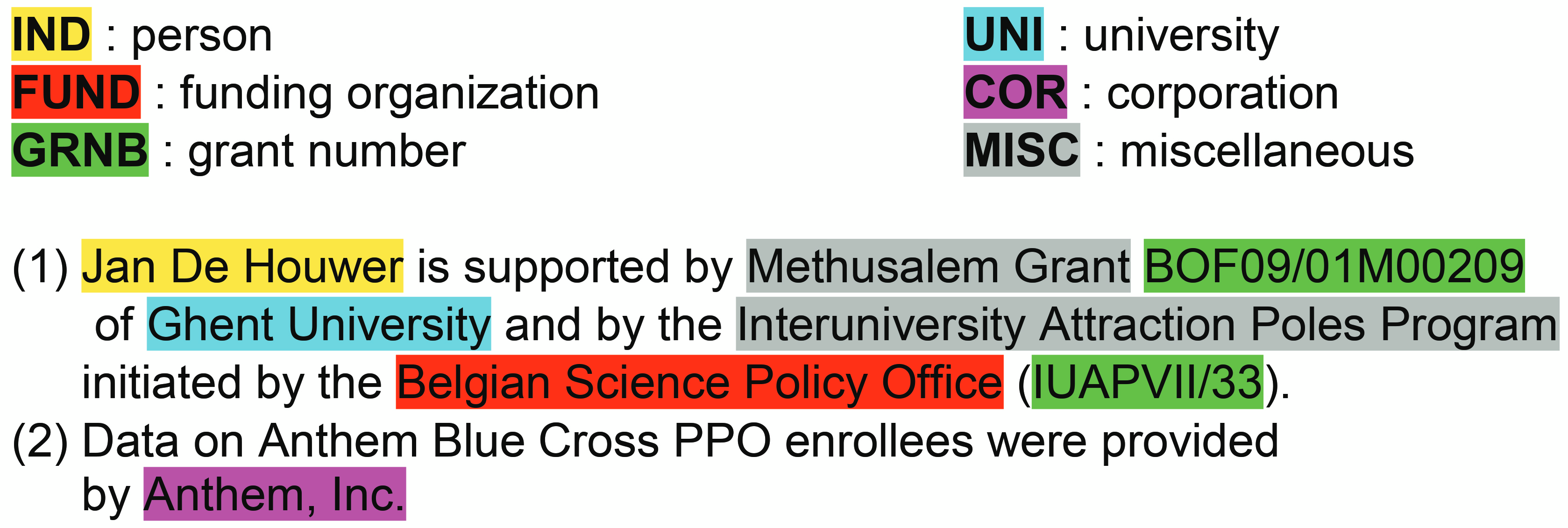}
  \caption{An example of acknowledged entities. Each entity type is marked with a distinct color.}
  \label{fig:example_ackn}
\end{figure}

In the present paper, different models for extraction and classification of acknowledged entities were evaluated. The choice of classification was inspired by Giles and Councill \cite[p.~17601]{giles_who_2004} classification: funding agencies (FUND), corporations (COR), universities (UNI), and individuals (IND). For our project, this classification was enhanced with the miscellaneous (MISC) and grant numbers (GRNB) categories. The GRNB category was adopted from WoS funding information indexing. The entities in the miscellaneous category could provide useful information but can not be ascribed to other categories, e.g., names of projects and names of conferences. Figure~\ref{fig:example_ackn} demonstrates an example of acknowledged entities of different types. To the best of our knowledge, Giles and Councill's classification is the only existing classification of acknowledged entities and therefore can be applied for the NER task. Other works on acknowledgement analysis focused on classification of acknowledgement texts. 

\subsection*{Training Data}
 
The Web of Science (WoS) database was used to harvest the training data (funding acknowledgments). From 2008 on, WoS started indexing information about funders and grants. WoS uses information from different funding reporting systems like researchfish\footnote{\url{https://mrc.ukri.org/funding/guidance-for-mrc-award-holders/researchfish/}}, Medline\footnote{\url{https://www.nlm.nih.gov/bsd/funding_support.html}} and others. 
As WoS contains millions of metadata records \cite{singh2021}, the data chosen for the present study was restricted by year and scientific domain. Records from four different scientific domains published from 2014 to 2019 were considered: two domains from the social sciences (sociology and economics) and oceanography and computer science. Different scientific domains were chosen, as previous works on acknowledgement analysis revealed the relations between scientific domain and types of acknowledged entities, i.e., acknowledged individuals are more characteristic of theoretical- and social-oriented domains, while information about technical and instrumental support are more common for the natural and life science domains \cite{diaz-faes_making_2017}. Only the WoS record types \textit{“article”} and \textit{“review”} published in a scientific journal in English were selected; then 1000 distinct acknowledgments texts were randomly gathered from this sample for the training dataset. Further different amount of sentences containing acknowledged entities were distributed into the differently-sized training corpora. Table~\ref{tab:ackn_corpus_count} demonstrates the amount of sentences in each set in the two corpora. We selected only sentences containing an acknowledged entity, regardless of the scientific domain. Table~\ref{tab:train_corpus_disc} contains the number of sentences from each scientific domain in the training corpora\footnote{The same article can belong to several scientific domains.}. 

\begin{table}[h!]
\begin{tabular}{ |p{1,2cm}||p{1,5cm}|p{1,5cm}|p{1,5cm}|p{1cm}| }
 \hline
 Corpus No.    & Training set (train)    & Test set (test) &   Validation set (dev) & \textbf{Total}\\
 \hline
 1 &   29  & 10   & 10 & \textbf{49}\\
 2 & 339 & 165 &  150 & \textbf{654}\\
 \hline
\end{tabular}
\caption{\label{tab:ackn_corpus_count} Number of sentences in the training corpora.}
\vspace{-8mm}
\end{table}

\begin{table}[h!]
\begin{tabular}{ |p{1cm}||p{1,8cm}|p{1,5cm}|p{1,2cm}|p{1,2cm}| }
 \hline
 Corpus No. & Oceanography & Economics & Social Sciences & Computer Science\\
 \hline
 1 &   13  & 3   & 20 & 16\\
 2 & 127 & 92 &  351 & 173\\
 \hline
\end{tabular}
\caption{\label{tab:train_corpus_disc} Number of \textbf{sentences} from each scientific domain in the training corpora.}
\vspace{-6mm}
\end{table}

\begin{figure}[h!]
\centering
  \includegraphics[width=0.5\textwidth]{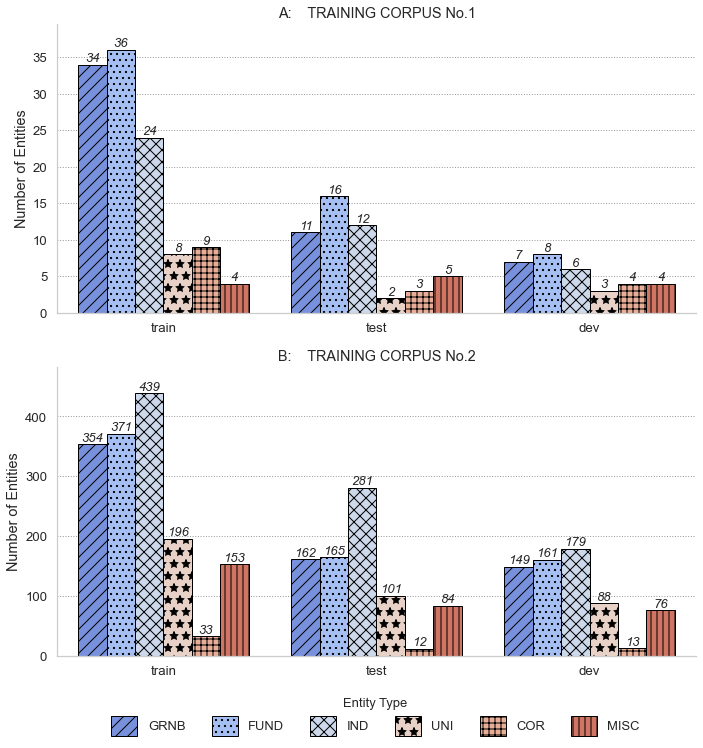}
  \caption{The distribution acknowledged entities of each type in the training corpora.}
  \label{fig:corpora_counts}
\end{figure}

Preliminary analysis of WoS data showed that WoS funding information indexing has several issues. The WoS includes only acknowledgements containing funding information; therefore, not every database entry has an acknowledgement, individuals are not included, and indexed funding organizations are not divided into different entity types like universities, corporations, etc. Therefore, existing indexing of funding organizations is incomplete. Furthermore, there is a disproportion between occurrences of acknowledged entities of different types. Thus, the most frequent entity types in the dataset with the training data are IND, FUND and GRNB, followed by UNI and MISC. COR is the most underrepresented category in the dataset. Consequently, there are different amounts of entities of different types in the training corpora (as Figure~\ref{fig:corpora_counts} demonstrates), which might have influenced the training results (see Section~\ref{sec:res}). 

The training corpus was annotated with six types of entities. As WoS already contains some indexed funding information, it was decided to develop a semi-automated approach for data annotation and use indexed information provided by WoS, therefore, grant numbers were adopted from the WoS indexing unaltered. 

Flair has a pre-trained 4-class NER Flair model (CoNLL-03)\footnote{\url{https://github.com/flairNLP/flair}}. The model is able to predict four tags: PER (person name), LOC ( location), ORG (organization name), and MISC (other name). As Flair showed adequate results in the extraction of names of individuals, it was decided to apply the pre-trained 4-class CoNLL-03 Flair model to the training dataset. Entities which fell into the PER category were added as the IND annotation to the training corpus. Furthermore, we noticed that some funding information was partially correctly extracted into the ORG and MISC categories. Therefore, WoS funding organization indexing and entities from the ORG and MISC categories were adopted and distinguished between three categories (FUND, COR and UNI) using regular expressions. Further, the automatic classification of entities was manually examined and reviewed. Mismatched categories, partially extracted entities, and not extracted entities were corrected. Acknowledged entities, which fall into the MISC category, were annotated manually.

\section{Default Training} \label{sec:def_train}
 
Training was performed using three default Flair training algorithms (described in Section~\ref{subsec:flair}): Flair Embeddings, Transformers, TARS. The default training was conducted with the recommended parameters for all algorithms, as Flair developers specifically ran various tests to find the best hyperparameters for the default models. Training was conducted with the small (corpus No.1) and bigger (corpus No.2) datasets. The small training corpus was mainly dedicated to test the TARS few-shot scenario. Additionally, we tested a zero-short scenario (without training data) for the TARS model.  

Flair Embeddings model uses the combination of static and contextual string embeddings. We applied GloVe \cite{pennington_glove_2014} as a static word-level embedding model.
Thus, in our case stacked embeddings comprise GloVe embeddings, forward contextual string embeddings and backwards contextual string embeddings.

For the Transformers the training was initiated with the RoBERTa model \cite{liu_roberta_2019}. For the present paper a fine-tuning approach was used. The fine-tuning procedure consisted of adding a linear layer to a transformer and retraining the entire network with a small learning rate. We used a standard approach, where only a linear classifier layer was added on the top of the transformer, as adding the additional CRF decoder between the transformer and linear classifier did not increase accuracy compared with this standard approach  \cite{schweter_flert_2020}.  

The TARS model requires labels to be defined in a natural language. Therefore we transformed our original coded labels into the natural language: FUND - “Funding  Agency”, IND - “Person”, COR - “Corporation”, GRNB - “Grant Number", UNI - “University”, and MISC - “Miscellaneous”.

\subsection{Results}\label{sec:res}

\begin{figure}[h!]
\centering
  \includegraphics[width=0.5\textwidth]{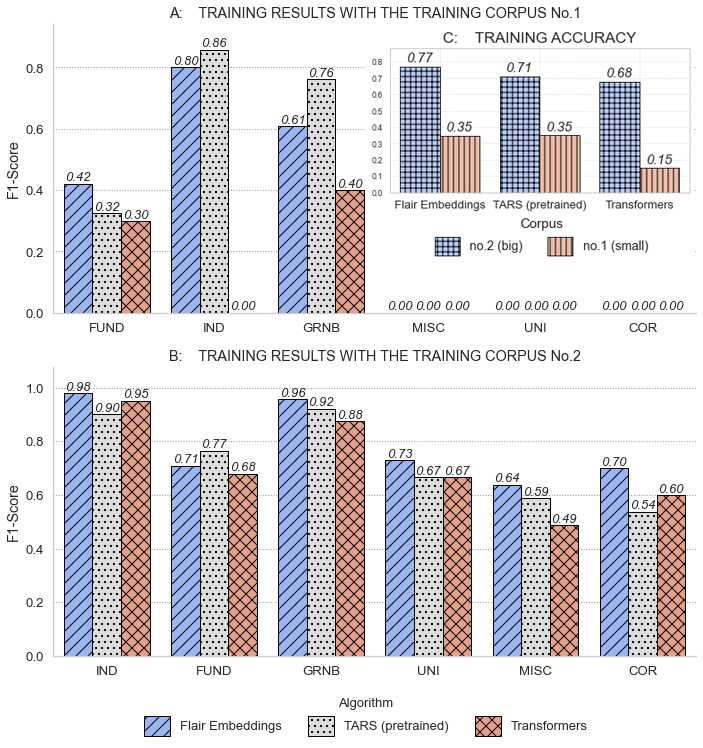}
  \caption{The training results with the training corpora No.1 (Figure A) and No.2 (Figure B). The figure comprises diagrams with the  F1-score (for each entity type) of the training with three algorithms. Figure C depicts the total accuracy of training algorithms.}
  \label{fig:prim_train_comb}
\end{figure}

Overall, training demonstrated mixed results. Figure~\ref{fig:prim_train_comb}-A depicts the training results with the corpus No.1. IND and GRNB showed adequate results by training with Flair Embeddings and TARS. IND was the best recognized entity by training with Flair Embeddings and TARS with a F1-score of 0.8 (Flair Embeddings) and 0.86 (TARS) respectively. The training with Transformers was not successful for IND with the F1-score of 0. Transformers overall showed to be a less efficient algorithm for training with the small dataset with the overall accuracy of 0.35 (Figure~\ref{fig:prim_train_comb}-C). FUND demonstrated not adequate results with F1-score less than 0.5 for all algorithms. Entity types MISC, UNI and COR showed the worst results with the F1-score equal to zero for all algorithms. Low accuracy for MISC, UNI and COR resulted in low overall accuracy for all algorithms. Overall training with the corpus No.1 showed insufficient results for all algorithms. Flair Embeddings and TARS, though, showed better accuracy in comparison with Transformers.
Further, training with the corpus No.2 was performed. Figure~\ref{fig:prim_train_comb}-B demonstrates training results with the corpus No.2. Similar to the training with corpus No.1, IND and GRNB are the best recognized categories. Best results for IND and GRNB demonstrated Flair Embeddings with the F1-score of 0.98 (IND) and 0.96 (GRNB). TARS achieved the best results for FUND with the F1-score of 0.77 against 0.71 for Flair Embeddings and 0.68 for Transformers. Miscellaneous demonstrated the worst accuracy for Flair Embeddings (0.64) and Transformers (0.49), while for TARS the worst accuracy lies in COR category with the F1-score of 0.54. Best result for UNI showed Flair Embeddings with the F1-score over 0.7. 

Training with corpus No.2 showed large improvement in training accuracy (Figure~\ref{fig:prim_train_comb}-C). Overall, Flair Embeddings was more accurate than other training algorithms, although training with TARS showed better results for the FUND category. Transformers showed the worst results during the training. 

Additionally a zero-shot approach was tested for the TARS model on corpus no.1. The model was able to successfully recognize individuals, but struggled with other categories, as Table~\ref{tab:tars_zero_acc} demonstrates. The total accuracy of the model comprises 0.23.

\begin{table}[h!]
\begin{tabular}{ |p{1cm}|p{1cm}|p{1cm}| p{1cm}|p{1cm}|p{1cm}| }
 \hline
 FUND & GRNB & IND &  UNI & COR & MISC \\
 \hline
 0.23  & 0.33   & 0.86 & 0 & 0& 0 \\
 \hline
\end{tabular}
\caption{\label{tab:tars_zero_acc} F1-Score of each category for the zero-shot learning.}
\vspace{-7mm}
\end{table}

\section{Additional training} \label{sec:add_train}

\begin{figure}[h!]
\centering
  \includegraphics[width=0.5\textwidth]{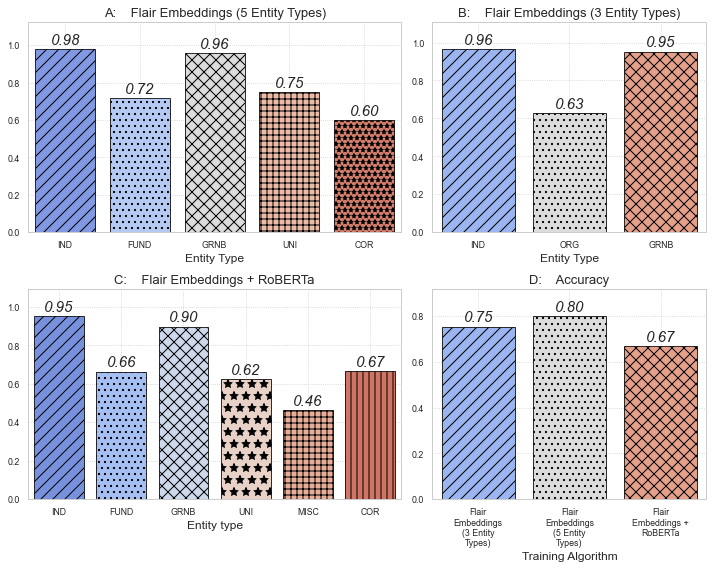}
  \caption{The results of the additional training. Figures A, B, C comprise diagrams with the F1-score of the additional training with three algorithms. Figure D represents the total accuracy of the training algorithms.}
  \label{fig:add_train_comb}
\end{figure}

Our first hypothesis to explain the pure model performance for the FUND, COR, MISC and UNI categories is that they are semantically close, which prevents successful recognition. Entities of these categories are often used in the same context. 
To examine this hypothesis, we conducted an experiment using Flair Embeddings with the dataset containing three types of entities: IND, GRNB and ORG. The MISC category was excluded from the training, as one of the aims of the present research is to extract information about acknowledged entities and the MISC category contains only additional information. The new ORG category was established, which includes a combination of entities from the FUND, COR and UNI categories. Training was performed with exactly the same parameters as the primary training with the Flair Embeddings model described in Section~\ref{sec:def_train}. Results of the training are represented in Figure~\ref{fig:add_train_comb}-B. The IND and GRNB still achieved high F1-scores of 0.96 (IND) and 0.95 (GRNB). Nevertheless, ORG gained only a F1-score of 0.64, which is worse than the previous results with six entity types. 

The UNI and COR categories, though, have distinct patterns. 
In this case, the low performance of the models for COR and UNI categories might be explained by the small size of the training sample containing these categories (see Figure~\ref{fig:corpora_counts}). Thus, the model was not able to identify patterns because of the lack of data.
Secondly, low results for FUND, COR, MISC and UNI categories might also lie in the nature of the miscellaneous category, as some entities that fall into this category are semantically very close to the FUND and COR categories. As a result, training without a MISC category might potentially show better performance.
To examine the second hypothesis, we conducted training with Flair Embeddings with a dataset excluding the MISC category, i.e., with five entity types. Training results are shown in Figure~\ref{fig:add_train_comb}-A. The results were quite similar to those achieved during the training with the dataset with six entity types. Improvement in overall accuracy (Figure~\ref{fig:add_train_comb}-D) (0.80 vs. the previous result of 0.77) could be explained by the fact that the MISC category was not present in this training and could not affect the overall accuracy with its low F1-score. 

Additionally, the problem might lie in the nature of training algorithms that were used. On the one hand, Flair developers claimed Transformers to be the most efficient algorithm \cite{schweter_flert_2020}. On the other, the stacked embeddings are an important feature of the Flair tool, as a combination of different embeddings might bring better results than their separate uses \cite{akbik_flair_2019}. Thus, the combination of the Transformer embeddings model with the contextual string embeddings might improve the model performance.
Thus, for the third additional training we combined contextual string embeddings with FLERT parameters. The training results are represented in Figure~\ref{fig:add_train_comb}-C. The proposed method showed no improvements compared to the results of the primary training with Transformers and worse performance compared with Flair Embeddings. 

\subsection*{Discussion}\label{disc_add_train}

The results of the additional training generally showed no improvement in the accuracy. On the contrary, training with the three entity types deteriorated the model performance. That might indicate that the model would make better predictions if the number of entity types is expanded. For example, the MISC category could be split into the following categories: names of projects, names of conferences, names of software and miscellaneous (for other possible information). Different subcategories could also be distinguished in the FUND category. Thus, this hypothesis requires further investigation. 

The model performance with Transformers and Flair Embeddings could also be improved by expanding the training corpus and adding more entries containing entities with the low recognition accuracy. Moreover using another transformer model such as SciBert \cite{beltagy-etal-2019-scibert} might increase the model performance.

\section{Conclusion}\label{sec:ret}
In this paper we evaluated different embedding models for the task of automatic extraction and classification of acknowledged entities from acknowledgement texts\footnote{The best model can be tested via an online demo available at: \url{https://colab.research.google.com/drive/1Wz4ae5c65VDWanY3Vo-fj__bFjn-loL4?usp=sharing}}. The annotation of the training corpora was the most challenging and time-consuming task of all data preparation procedures. Therefore, a semi-automated approach was used to help significantly accelerate the procedure. 
Flair Embeddings showed the best accuracy in the training with the bigger corpus and the fastest training time in comparison with the other models; thus, it is recommended to further use the Flair Embeddings model for the recognition of acknowledged entities. Expanding the size of a training corpus massively increased the accuracy of all training algorithms. 
Main limitations of the study were the small sizes and just one annotator of the training corpora.  

\subsection*{Acknowledgement}

The work was funded by German Centre for Higher Education Research and Science Studies (DZHW) via the project "Mining Acknowledgement Texts in Web of Science (MinAck)"\footnote{\url{https://kalawinka.github.io/minack/}}. Access to the WoS data was granted via the Competence Centre for Bibliometrics\footnote{\url{https://www.bibliometrie.info/en/index.php?id=home }}. Data access was funded by BMBF (Federal Ministry of Education and Research, Germany) under grant number 01PQ17001.

\clearpage
\bibliographystyle{ACM-Reference-Format}
\bibliography{refs}

\end{document}